\documentclass[runningheads]{llncs}
\usepackage[T1]{fontenc}
\usepackage{graphicx}
\usepackage{multirow}
\usepackage{booktabs} 
\usepackage[table]{xcolor}
\usepackage{subcaption}
\newcommand{\red}[1]{\textcolor{black}{#1}}
\begin{document}
\title{Towards Khmer Scene Document Layout Detection}
%
%

\author{Marry Kong\inst{1} \and Rina Buoy\inst{1,2} \and Sovisal Chenda\inst{1} \and Nguonly Taing\inst{1} \and Masakazu Iwamura\inst{2} \and Koichi Kise\inst{2}}
%
\authorrunning{Kong et al.}
%
\institute{Techo Startup Center, Ministry of Economy and Finance, Cambodia \and Osaka Metropolitan University, Japan}
\maketitle              
\begin{abstract}
While document layout analysis for Latin scripts has advanced significantly, driven by the advent of large multimodal models (LMMs), progress for the Khmer language remains constrained because of the scarcity of annotated training data. This gap is particularly acute for scene documents, where perspective distortions and complex backgrounds challenge traditional methods. \red{Given the structural complexities of Khmer script, such as diacritics and multi-layer character stacking, existing Latin-based layout analysis models fail to accurately delineate semantic layout units, particularly for dense text regions (e.g., list items).} In this paper, we present the first comprehensive study on Khmer scene document layout detection. We contribute a novel framework comprising three key elements: (1) a robust training and benchmarking dataset specifically for Khmer scene layouts; (2) an open-source document augmentation tool capable of synthesizing realistic scene documents to scale training data; and (3) layout detection baselines utilizing YOLO-based architectures with oriented bounding boxes (OBB) to handle geometric distortions.  To foster further research in the Khmer document analysis and recognition (DAR) community, we release our models, code, and datasets in this gated repository\footnote{in review}.

\end{abstract}
\section{Introduction}\label{intro}

In the field of document analysis and recognition (DAR), document layout analysis (DLA) is a key step in OCR-based document extraction and parsing pipelines, as it identifies semantic regions within a given page image. These semantic regions include, for example, text blocks, figures, tables, headers, and other structural components~\cite{markewich2022segmentation}. Over the past decade, DLA methods, along with training and evaluation datasets for Latin-based scripts and other high-resource languages, have advanced significantly, from early semantic segmentation and object-detection-based approaches to recent OCR-free multimodal and vision language models.

Nonetheless, DLA for complex low-resource languages, such as Khmer, remains underexplored~\cite{thuon2025kh}. This is largely due to the fact that Khmer is a complex abugida writing system featuring multi-layer character stacking and the absence of explicit word boundaries~\cite{buoy2023toward}. \red{As a result, existing Latin-based layout analysis models fail to accurately delineate semantic layout units, particularly for dense text regions (e.g., text blocks, list items).} Moreover, there are very few annotated datasets (e.g., KH-FUNSD~\cite{thuon2025kh} for scanned Khmer business forms, and KhmerST~\cite{nom2024khmerst} and WildKhmerST~\cite{nom2025wildkhmerst} for scene text detection and recognition), as well as a scarcity of pre-trained models for Khmer DLA, particularly for scene document images. Such images are often captured by smartphone cameras in natural environments, leading to severe distortions and complex backgrounds. In the Cambodian context, scene images are more common than scanned documents due to accessibility and convenience.

At the time of writing, there are no published Khmer-native DLA models specifically designed to segment Khmer documents into semantic layout units. Existing multilingual DLA models and frameworks, such as Surya-OCR~\cite{paruchuri2025surya}, PaddleOCR~\cite{cui2025paddleocr30technicalreport}, Docling~\cite{Docling}, and DeepSeek-OCR~\cite{wei2025deepseek,wei2026deepseek}, provide only limited support for Khmer, and primarily for scanned document images without any significant perspective transforms and distortions.

Thus, to address the challenges associated with Khmer DLA, which is crucial for Khmer document digitization, we present the first comprehensive study on Khmer scene document layout detection. Our work includes the construction of the first training and benchmark datasets for Khmer scene documents, the development of a scene document layout augmentation tool, and the training of baseline Khmer DLA models. \red{One of the key distinctions of our dataset is that it includes both standard and modern Khmer document formats, such as books, press releases, PowerPoint presentations, and complex infographics. Similarly, while existing standard scene text recognition augmentation tools (e.g., SynthTiger~\cite{yim2021synthtiger}) transform only characters, our compositional layout augmentation tool must transform both characters and layouts (i.e., bounding boxes) proportionally. This requirement necessitates the implementation of a more complex linear and non-linear  transformation pipeline. In addition, rather than using all augmented layout images, strict manual verification by human annotators is required to keep only the valid images with the structural integrity of the layouts.  }

Our contributions can be summarized as follows:

\begin{enumerate}
\item We construct the first Khmer scene document layout dataset for training and evaluation, which, at the time of writing, is the largest single training and evaluation dataset for the Khmer script for the layout analysis task.

\item To scale the training data and simulate real Khmer scene documents, we develop a compositional layout augmentation tool capable of synthesizing realistic scene document images.

\item Lastly, we train and evaluate the first Khmer scene document layout detection baselines using YOLO-based architectures using our newly-constructed training and evaluation data.

\end{enumerate}


\section{Related Work}\label{relatedwork}
\subsection{Latin Document Layout Analysis }\label{relatedwork_latin}

Early approaches to document layout analysis and detection, particularly for Latin-based scripts, formulated this problem as either object detection or semantic segmentation. Shen et al.~\cite{Shen_2020_CVPR_Workshops} applied object detection methods to layout analysis and detection on historical Japanese documents using Faster R-CNN~\cite{ren2015faster} and RetinaNet~\cite{lin2017focal}. Similarly, Zhao et al.~\cite{zhao2024doclayout} improved the YOLOv10~\cite{ultralytics_ultralytics_github} backbone and trained it for layout analysis and detection on a large-scale synthetic dataset, DocSynth300K. In contrast, Markewich et al.~\cite{markewich2022segmentation} formulated layout analysis as a semantic segmentation task, which is a pixel labeling task.

With the advent of Transformers networks~\cite{vaswani_attention_2017}, particularly Vision Transformers (ViT)~\cite{dosovitskiy2020image}, Transformers-based approaches to DLA have become increasingly popular. Li et al.~\cite{Li2022} proposed the Document Image Transformers (DiT), which was pre-trained on large-scale unlabeled document images and subsequently fine-tuned for a range of document AI tasks, including document layout analysis and beyond. Beyond purely visual representations, multimodal frameworks such as LayoutLM and its variants~\cite{xu2020layoutlm,xu2021layoutlmv2,huang2022layoutlmv3} incorporate visual, textual (often derived from OCR results), and geometric cues to further enhance detection and recognition performance.

Recently, vision-language models (VLMs), such as InternVL3.5~\cite{wang2025internvl3}, Qwen3-VL~\cite{bai2025qwen3}, and DeepSeek-OCR and its variants~\cite{wei2025deepseek,wei2026deepseek}, have dramatically transformed the field of DLA. Many modern VLMs can be instructed via natural language prompts to perform a wide range of vision-related tasks, including DLA and beyond, within a single unified model. Nonetheless, many of these models are optimized for high-resource languages and therefore underperform on low-resource scripts, such as Khmer. 

\subsection{Khmer Document Layout Analysis}\label{relatedwork_latin}

\begin{figure}[t]
\centering

\centering
    {\includegraphics[width=\hsize]{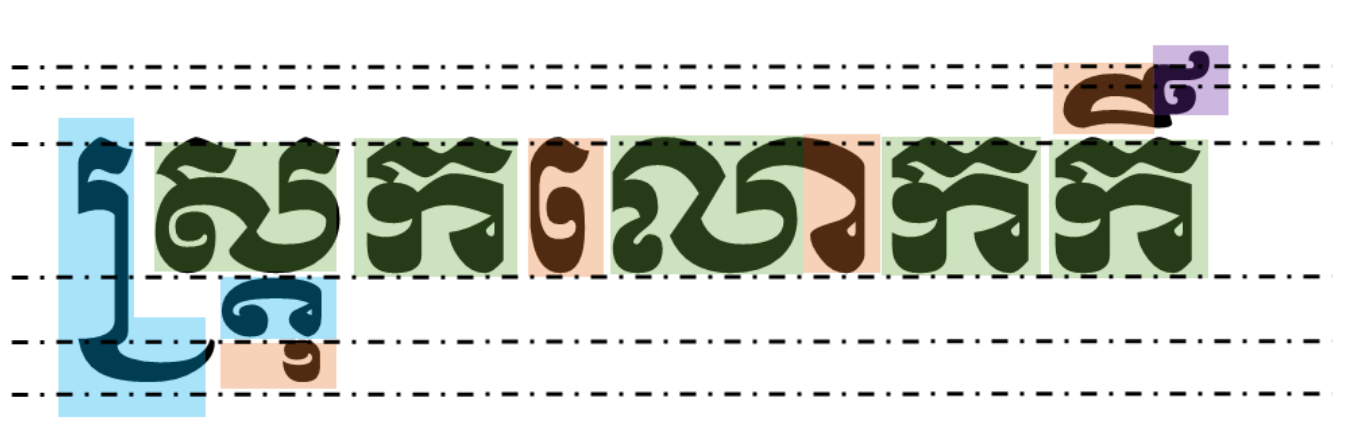}}
    \caption{Sample Khmer text layout (with permission~\cite{buoy2025addressing}.)  Blue: consonant subscript. Green: base consonant. Orange: dependent vowel. Purple: diacritic. Best viewed in color. }

\label{khmer_text_layout}
\end{figure}

\red{Khmer is a complex abugida writing system with a rich inventory of characters, including consonants (base and subscript forms), dependent and independent vowels, and diacritics~\cite{buoy2023toward}, as illustrated in Figure~\ref{khmer_text_layout}. The script is written from left to right, with spaces used optionally. Characters often form multi-level stacked clusters by attaching subscripts to base consonants, producing intricate ligature structures~\cite{buoy2023toward}. These complex spatial arrangements not only complicate character recognition but also pose significant challenges for layout analysis, as text regions exhibit highly variable shapes, densities, and reading orders.}

The development of Khmer DLA and related research remains relatively limited compared to that for Latin scripts. KH-FUNSD~\cite{thuon2025kh} is the first publicly available Khmer layout dataset for form document understanding, featuring hierarchical FUNSD-style annotations~\cite{jaume2019funsd}. The dataset includes receipts, invoices, and quotations, and is split into training (111), validation (23), and test (24) sets. It has been used to train YOLO~\cite{ultralytics_ultralytics_github}-, DETR~\cite{carion2020end}-, and LayoutLM~\cite{xu2020layoutlm}-based baseline models. Nom et al.~\cite{nom2024khmerst} introduced KhmerST, the first Khmer scene textline detection and recognition dataset, which contains 1,544 real-world images (997 indoor, 547 outdoor) with line-level text annotations. WildKhmerST~\cite{nom2025wildkhmerst} is an extended version of KhmerST, comprising 29,601 annotated textlines from 10,000 images captured in diverse public locations across Cambodia, including streets, signboards, supermarkets, and commercial areas.

In summary, DLA for Khmer and other low-resource languages remains relatively unexplored due to limited research attention and, importantly, the lack of large-scale annotated datasets. Thus, we aim to address this challenges in this study.

\section{Methodology}\label{methodology}

\begin{figure}[t]
\centering
\includegraphics[width=\textwidth]{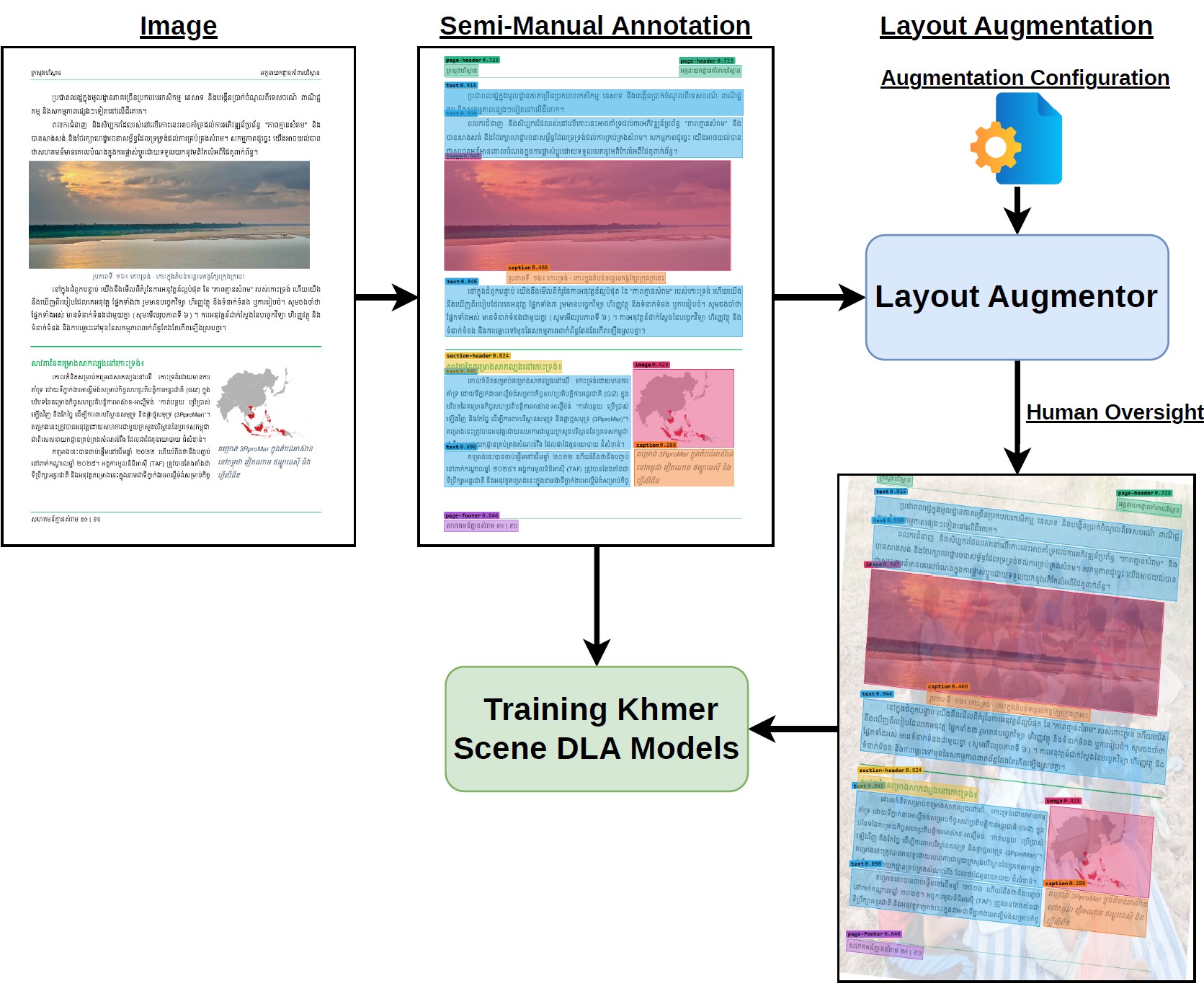}
\caption{\red{The overall methodology of this study.}} \label{overall_method}
\end{figure}

Figure~\ref{overall_method} illustrates the overall methodology of this study, highlighting three key components: Khmer DLA data construction, layout augmentation using a layout augmentor, and model training with both the original and augmented DLA data.

\subsection{Dataset Construction}
\begin{table}[t]
\centering
\caption{Layout label scheme: labels and descriptions.}
\label{tab:surya_labels}
\begin{tabular}{ll}
\toprule
\textbf{Label} & \textbf{Description} \\
\midrule
caption & Descriptive text associated with figures or tables \\
equation-block & Standalone mathematical expressions or displayed equations \\
figure & Diagrams, charts, or visual elements conveying structured information \\
footnote & Bottom-page explanatory or reference notes \\
form & Structured layouts containing fillable fields (e.g., invoices, surveys) \\
image & Embedded or decorative images not classified as figures \\
list-item & Individual items in bulleted or numbered lists \\
page-footer & Bottom margin content such as page numbers or copyright notices \\
page-header & Top margin content such as titles, running headers, or logos \\
section-header & Titles indicating section or subsection boundaries \\
table & Structured tabular content arranged in rows and columns \\
table-of-contents & Structured overview of document organization and sections \\
text & Main body paragraph text of the document \\
\bottomrule
\end{tabular}
\end{table}

\begin{table}
\centering
\caption{Distribution of layout labels in the training and validation sets.}
\label{tab:label_distribution_train_val}
\begin{tabular}{lrrr}
\toprule
\textbf{Label} & \textbf{Train} & \textbf{Validation} & \textbf{Total} \\
\midrule
caption & 3{,}012 & 791 & 3{,}803 \\
code-block & 2 & 0 & 2 \\
equation-block & 37 & 3 & 40 \\
figure & 1{,}052 & 193 & 1{,}245 \\
footnote & 946 & 120 & 1{,}066 \\
form & 233 & 0 & 233 \\
image & 3{,}289 & 110 & 3{,}399 \\
list-item & 13{,}898 & 1{,}522 & 15{,}420 \\
page-footer & 8{,}794 & 1{,}096 & 9{,}890 \\
page-header & 1{,}853 & 3 & 1{,}856 \\
section-header & 8{,}424 & 1{,}503 & 9{,}927 \\
table & 3{,}256 & 526 & 3{,}782 \\
table-of-contents & 236 & 73 & 309 \\
text & 17{,}630 & 2{,}194 & 19{,}824 \\
\midrule
\textbf{Total} & \textbf{62{,}662} & \textbf{9{,}134} & \textbf{71{,}796} \\
\bottomrule
\end{tabular}
\end{table}

We collected publicly available PDF documents from the Open Development Cambodia data sharing portal\footnote{https://opendevelopmentcambodia.net/km/}
 and others. The collected documents are scanned and cover diverse domains, ranging from technology to culture, and include various types, from books to infographics and PowerPoint presentations. In this study, we follow the label scheme of Surya-OCR due to its simplicity for human data annotators. The labels and their corresponding descriptions are provided in Table~\ref{tab:surya_labels}. Our data annotation strategy comprises two stages as follows:

\begin{itemize}
    \item \textbf{Screening} In this stage, we use Surya-OCR to perform document layout analysis. Although the multilingual Surya-OCR is not designed to optimally support the Khmer language, it can capture language-agnostic labels, such as figures, images, tables, and tables of contents, which can significantly reduce the workload for human annotators. Nonetheless, substantial labeling errors, particularly for script-dependent labels such as section headers, list items, and captions,   can still occur because Surya-OCR is not optimized for Khmer script and layouts. For example, captions and list items are sometimes mislabeled as text, while certain Khmer tables and footnotes may not be detected at all.

    \item  \textbf{Human Curation}: In this stage, human annotators are provided with the layout label scheme in Table~\ref{tab:surya_labels} and are tasked with adding missing labels and correcting any mislabeled regions. For this labeling and annotation task, we use the LabelMe\footnote{https://github.com/wkentaro/labelme}
 tool with polygon-based annotations.  
\end{itemize}

\begin{figure}[t]
\centering

\begin{subfigure}[b]{\textwidth}
    \centering
    {\includegraphics[width=\textwidth]{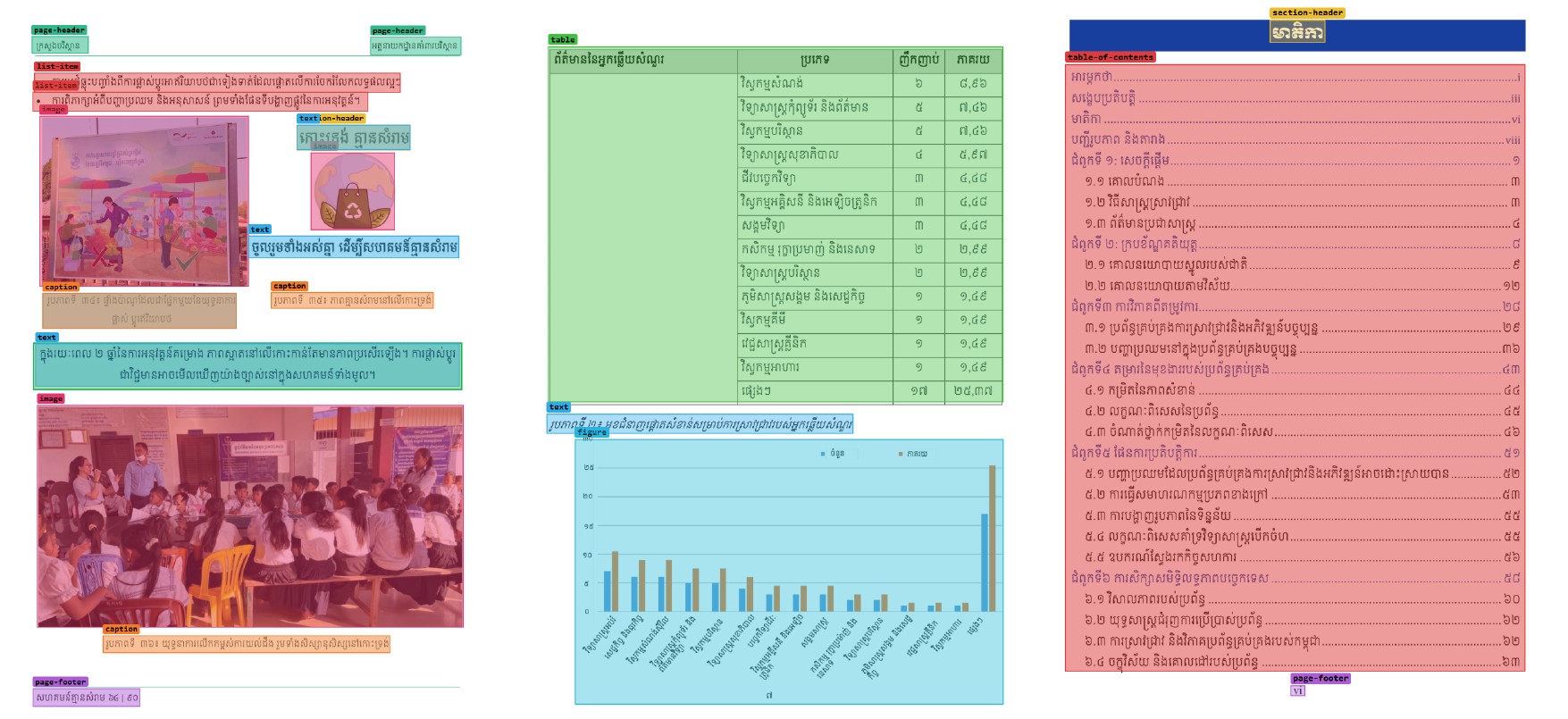}}
    \caption{Images with captions, table, figure, table of content, and others}

\end{subfigure}
\hfill
\begin{subfigure}[b]{\textwidth}
    \centering
    {\includegraphics[width=\textwidth]{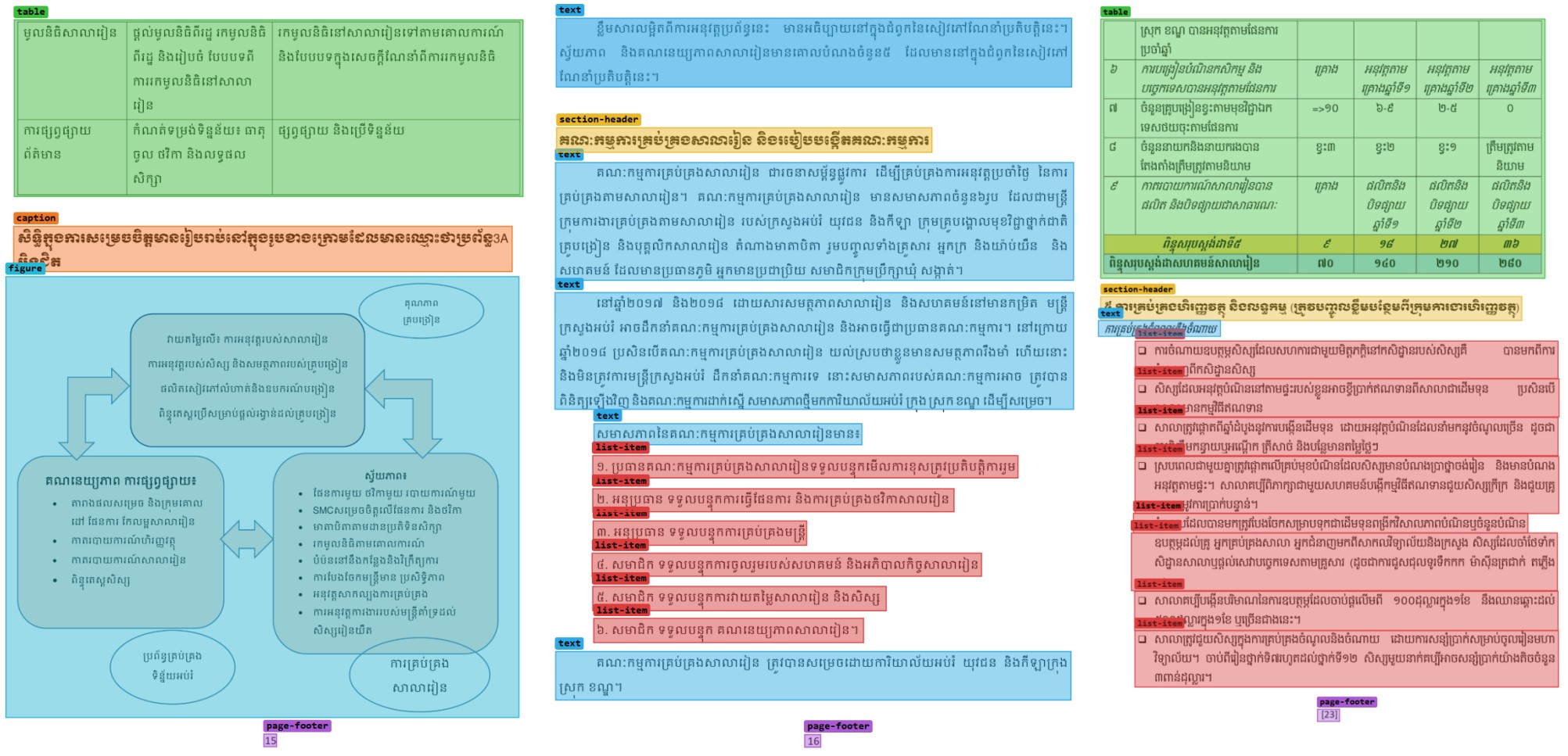}}
    \caption{Figure, tables, list-items and others.}

\end{subfigure}

\caption{A few sample annotated images with human curations.} \label{samples_pages}
\end{figure}

We converted the PDF documents into a total of 8,990 image pages and applied the annotation strategy described above. The final curated dataset was split into training (7,818 pages) and evaluation (1,178 pages) sets, which are significantly larger than KH-FUNSD~\cite{thuon2025kh} in terms of both scale and scope. The label distributions for the training and evaluation datasets are provided in Table~\ref{tab:label_distribution_train_val}. The distributions indicate that code blocks and equation blocks are underrepresented, as the source documents are general-domain rather than focused on mathematics or programming. A few sample annotated images with human curation are shown in Figure~\ref{samples_pages}.

\subsection{Layout Augmentation}
\begin{figure}[t]
\centering

\begin{subfigure}[b]{0.45\textwidth}
    \centering
    {\includegraphics[width=\textwidth]{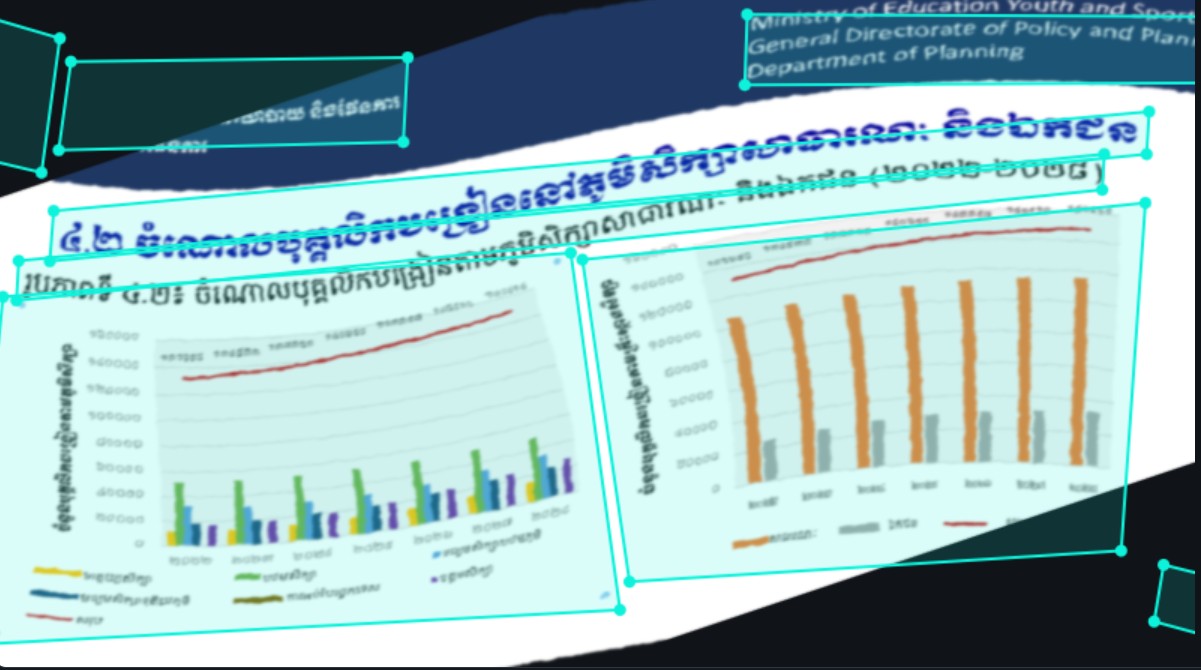}}

\end{subfigure}
\hfill
\begin{subfigure}[b]{0.49\textwidth}
    \centering
    {\includegraphics[width=\textwidth]{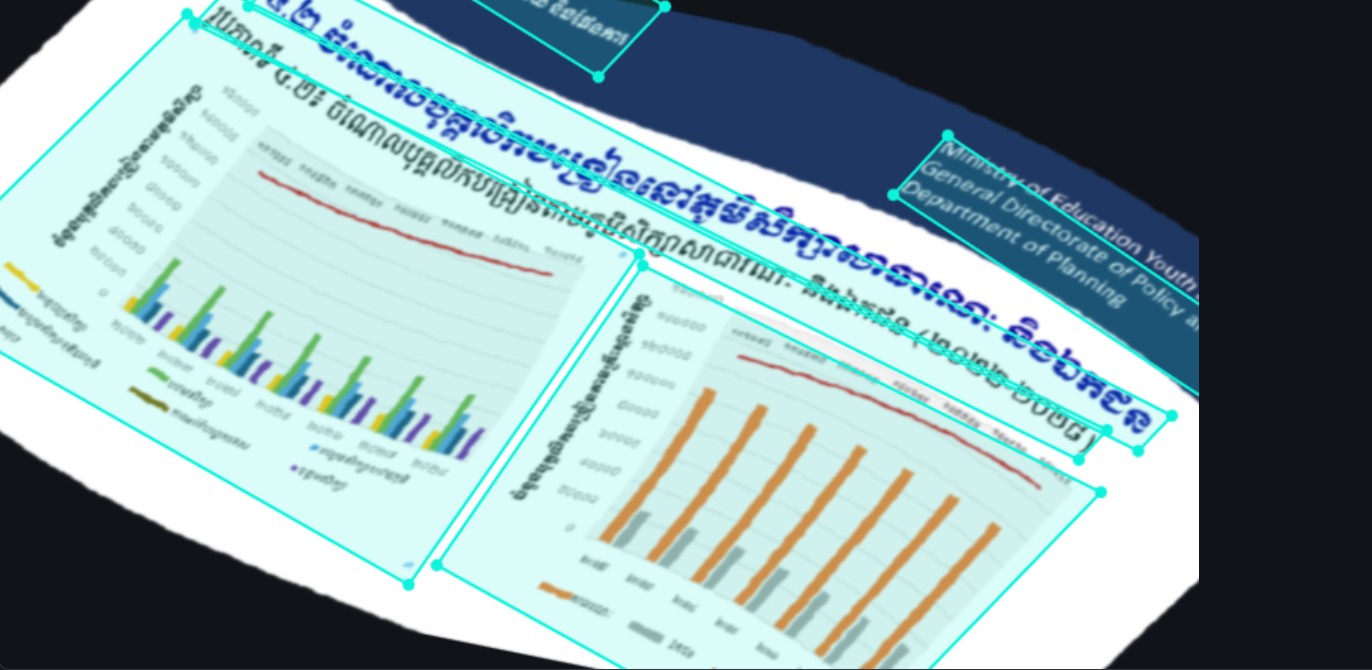}}
 \end{subfigure}   
\caption{\red{A few sample extremely augmented cases with corrupted bounding boxes.}} \label{samples_pages_augmented_bad}
\end{figure}
The annotated document images, converted from the PDF sources, do not contain scene distortions or deformations. In real-world scenarios, however, document images are often captured in the wild using smartphone cameras and exhibit complex distortions and deformations. To address this, we developed a document layout augmentor. The augmentor employs a composition-based transformation pipeline, configured via a JSON file.

\red{Specifically, we use a two-stage augmentation pipeline that ensures pixel-accurate correspondence between transformed images and their annotations (i.e., bounding boxes). The pipeline applies (1) non-linear deformation augmentations followed by (2) affine geometric transformations, with all transformations applied identically to both image pixels and annotation coordinates.}

\red{Non-linear deformations are applied first to introduce realistic, non-affine variations while maintaining local structure. The following deformation types are available:}
\red{
\begin{itemize}
\item Elastic deformation: Perlin noise-based displacement fields that simulate organic material deformations.
\item Grid distortion: Sinusoidal displacement creating grid-like warping patterns.
\item Barrel/pincushion distortion: Radial lens distortion from the image center.
\item Wave distortion: Periodic sinusoidal waves applied in both directions.
\item Radial swirl distortion: Rotation around the center proportional to distance, creating swirl effects.
\end{itemize}}

\red{After non-linear deformations, affine transformations are applied via a single transformation matrix. This ensures computational efficiency and maintains straight lines and parallelism. The complete affine transformation is represented as a $3\times3$ homogeneous transformation matrix, composed through sequential matrix multiplication. Transformations are applied from right to left, with the image center as the transformation origin. The composition includes:}
\red{
\begin{itemize}
\item Translation to origin: Move image center to coordinate origin.
\item Flip operations: Horizontal and/or vertical axis reflection.
\item Scale: Independent $x$ and $y$ scaling factors.
\item Shear: Skewing along $x$ and $y$ axes.
\item Rotation: Rotation by angle $\phi$ around the center.
\item Perspective (affine approximation): Simplified perspective using affine approximation.
\item Translation to center: Move image center back to original position.
\end{itemize}}
\red{
Annotation shapes are transformed using the following approach:}
\red{
\begin{itemize}
\item Rectangle to polygon conversion: Convert two-point bounding boxes to four-corner polygons.
\item Shape transformation: Apply the complete transformation pipeline to each annotation point.
\end{itemize}}

\begin{figure}[t]
\centering

\begin{subfigure}[b]{\textwidth}
    \centering
    {\includegraphics[width=\textwidth]{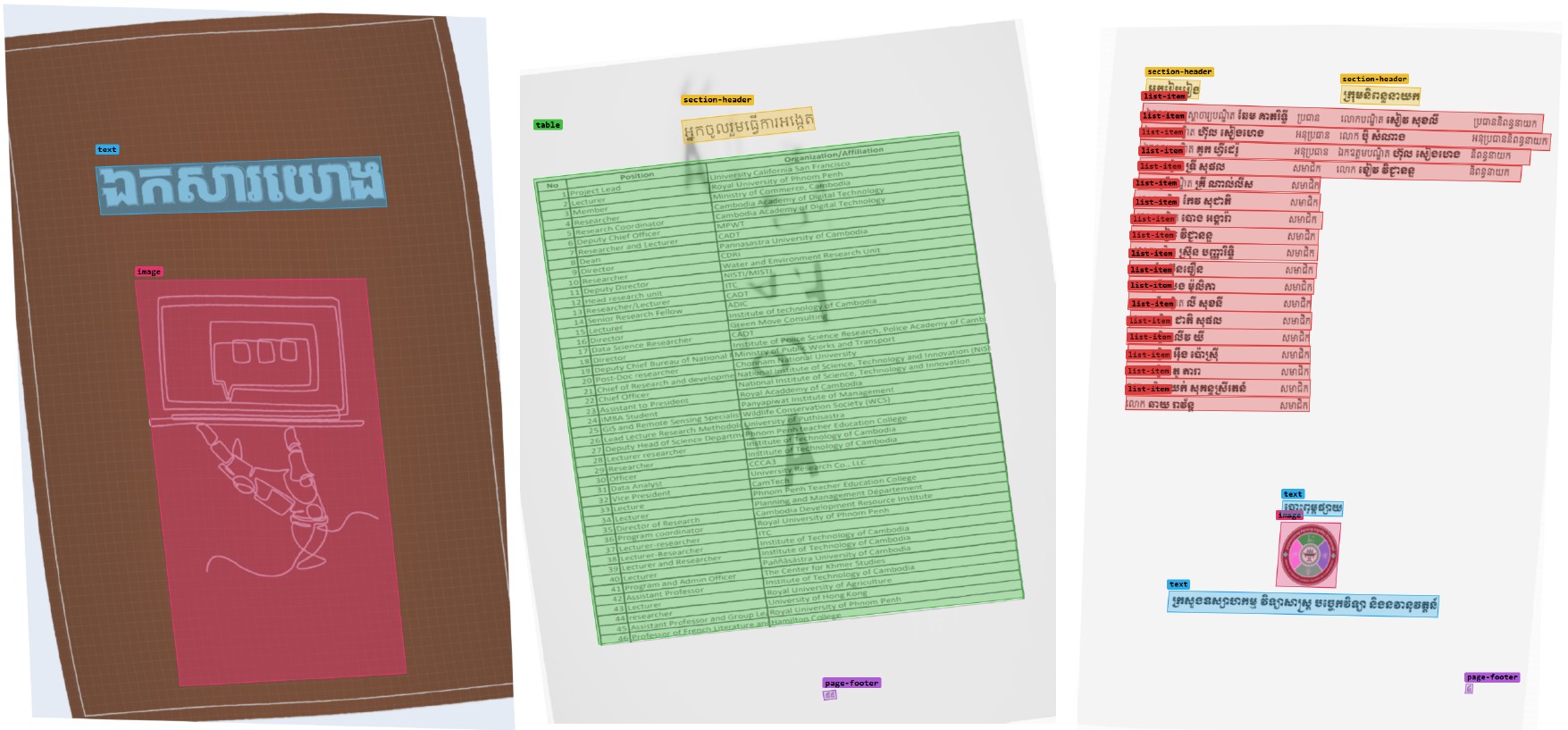}}
    \caption{Texts, image, table, list-items, and others.}

\end{subfigure}
\hfill
\begin{subfigure}[b]{\textwidth}
    \centering
    {\includegraphics[width=\textwidth]{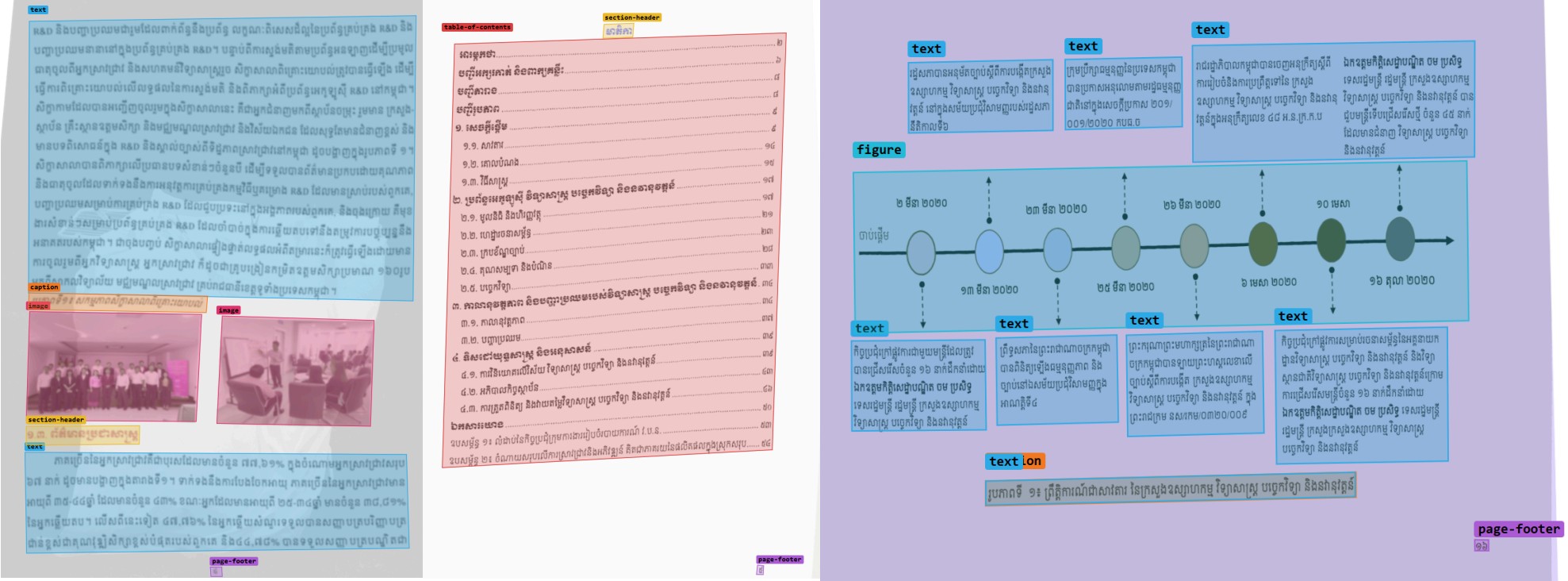}}
    \caption{Images, figure, table of content, and others.}
 \end{subfigure}   
\caption{A few sample augmented images with augmented annotations.} \label{samples_pages_augmented}
\end{figure}

\red{The key property of our pipeline is that the same transformation functions are applied to both image pixels and annotation coordinates. This ensures pixel-accurate correspondence between augmented images and their annotations}. Starting from the human-curated document images and annotations, we generated augmentation configuration files for the layout augmentor by randomizing each parameter within its realistic range. The resulting augmented document images were then reviewed by human annotators to remove any images with unrealistically extreme augmentations, \red{as shown in Figure~\ref{samples_pages_augmented_bad}}. After this curation process, we obtained an additional 2,258 labeled images with correctly transformed polygon annotations.  A few sample augmented images with augmented annotations are shown in Figure~\ref{samples_pages_augmented}.

\subsection{Training Khmer Scene DLA Models}

Many standard object detection approaches for general DLA rely on horizontally aligned bounding boxes (i.e., rectangles), which are suitable for non-deformed or well-aligned document images. However, restricting annotations to only horizontally aligned bounding boxes is impractical for scene document images, which are common in the Khmer context.

To accurately detect oriented bounding boxes (OBBs), we trained YOLO-based architectures~\cite{ultralytics_ultralytics_github} specifically designed for OBB detection. The key difference is that, in addition to the standard outputs, the model predicts an extra parameter representing the angle of the bounding box. The YOLO-based models used in this study include YOLO11-OBB, YOLO12-OBB, and YOLO26-OBB, each available in small, medium, large, and extra-large variants. The comparison in terms of model size, parameters, floating point operations (FLOPs), runtime, and frame per second (FPS) is given in Table~\ref{tab:yolo_model_comparison}.

\begin{table*}[t]
\centering
\caption{Comparison of YOLO11, YOLO12, and YOLO26 model variants with 640$\times$640 input resolution.}
\label{tab:yolo_model_comparison}
\begin{tabular}{lccccccc}
\toprule
\textbf{Model} & \textbf{Base} & \textbf{Size} & \textbf{File (MB)} & \textbf{Params (M)} & \textbf{FLOPs (G)} & \textbf{Runtime (ms)} & \textbf{FPS} \\
\midrule
YOLO11s & 11 & s & 18.93 & 9.72 & 11.28 & 73.13 & 13.67 \\
YOLO11m & 11 & m & 40.37 & 20.91 & 35.97 & 133.78 & 7.48 \\
YOLO11l & 11 & l & 50.55 & 26.17 & 45.51 & 180.73 & 5.53 \\
YOLO11x & 11 & x & 112.84 & 58.80 & 101.92 & 275.18 & 3.63 \\
\midrule
YOLO12s & 12 & s & 18.70 & 9.54 & 11.27 & 164.22 & 6.09 \\
YOLO12m & 12 & m & 40.61 & 21.00 & 35.74 & 263.33 & 3.80 \\
YOLO12l & 12 & l & 52.78 & 27.25 & 46.58 & 431.02 & 2.32 \\
YOLO12x & 12 & x & 117.34 & 61.04 & 104.10 & 702.05 & 1.42 \\
\midrule
YOLO26s & 26 & s & 20.59 & 10.53 & 12.26 & 79.72 & 12.54 \\
YOLO26m & 26 & m & 45.36 & 23.49 & 41.11 & 139.45 & 7.17 \\
YOLO26l & 26 & l & 53.91 & 27.89 & 50.31 & 176.27 & 5.67 \\
YOLO26x & 26 & x & 120.28 & 62.66 & 112.64 & 277.29 & 3.61 \\
\bottomrule
\end{tabular}
\end{table*}

\section{Experimental Setup}\label{experiments} 
We employed YOLO11, YOLO12, and YOLO26 with OBB detection capabilities for the scene document detection task. For each YOLO architecture, we experimented with four model variants: small (s), medium (m), large (l), and extra-large (x). The models were initialized with their respective pre-trained weights and fine-tuned on both our human-curated and augmented datasets to detect the 13 document layout element types described earlier in Table~\ref{tab:surya_labels}.

For data preprocessing, we converted annotations from polygon coordinates to the OBB format. Ground-truth bounding boxes were represented using four points in clockwise order starting from the top-left corner. All coordinates were normalized to the $[0, 1]$ range relative to the image dimensions. The models were trained for 100 epochs with a batch size of eight and input image size of 640×640 pixels. Training was accelerated using a single NVIDIA RTX 6000 GPU with four parallel data-loading workers.

Stochastic gradient descent (SGD) was used as the optimizer, with a momentum of 0.937 and a weight decay of 0.0005. The learning rate followed a cosine annealing schedule, starting from an initial learning rate ($lr_0$) of 0.01 and decaying to a final learning rate ($lr_f$) of 0.01 over the course of training.

\section{Results and Discussion}\label{experiments} 
\subsection{Layout Detection Performance}\label{experiments} 

\begin{table*}[t]
\centering
\caption{Overall layout detection performance of the baseline YOLO-based models on the evaluation set. \textbf{Bold}: \emph{best}. \textbf{Italic}: \emph{second best}.}
\label{tab:layout_metrics}
\begin{tabular}{lccccc}
\toprule
\textbf{Model} & \textbf{Precision} & \textbf{Recall} & \textbf{mAP@0.5} & \textbf{mAP@0.75} & \textbf{mAP@0.5:0.95} \\
\midrule
YOLO11s & 0.9547 & 0.9217 & 0.9532 & 0.9493 & 0.9356 \\
YOLO11m & 0.9813 & 0.9258 & 0.9571 & 0.9513 & 0.9424 \\
YOLO11l & 0.9788 & 0.9228 & 0.9565 & 0.9523 & 0.9360 \\
YOLO11x & 0.9824 & 0.9011 & 0.9461 & 0.9435 & 0.9306 \\
\midrule
YOLO12s & \textbf{0.9843} & 0.9213 & \emph{0.9603} & 0.9568 & 0.9439 \\  
YOLO12m & 0.9783 & \textbf{0.9365} & \textbf{0.9621} & \textbf{0.9588} & 0.9440 \\
YOLO12l & 0.9758 & 0.9227 & 0.9588 & 0.9565 & \emph{0.9479} \\
YOLO12x & 0.9528 & \emph{0.9364} & 0.9594 & \emph{0.9582} & \textbf{0.9502} \\
\midrule
YOLO26s & 0.9702 & 0.9127 & 0.9496 & 0.9454 & 0.9338 \\
YOLO26m & 0.9764 & 0.9278 & 0.9587 & 0.9524 & 0.9402 \\
YOLO26l & 0.9469 & 0.9234 & 0.9507 & 0.9454 & 0.9363 \\
YOLO26x & \emph{0.9834} & 0.9128 & 0.9528 & 0.9506 & 0.9412 \\

\bottomrule
\end{tabular}
\end{table*}

\begin{table*}[t]
\centering
\caption{Per-class layout detection performance of the YOLO12x model on the evaluation set.  \textbf{Italic}:\emph{second best}.}
\label{tab:per_class_metrics}
\begin{tabular}{lrrrrrr}
\toprule
Class & Images & Instances & P & R & mAP@0.5 & mAP@0.5:0.95 \\
\midrule
caption & 513 & 797 & 0.972 & 0.967 & 0.978 & 0.954 \\
equation-block & 2 & 3 & \textbf{1.000} & \textbf{1.000} & \textbf{0.995} & 0.973 \\
figure & 124 & 160 & 0.994 & 0.975 & 0.987 & 0.986 \\
footnote & 86 & 122 & 0.980 & 0.803 & 0.898 & 0.882 \\
image & 89 & 144 & 0.897 & 0.972 & 0.964 & 0.958 \\
list-item & 258 & 1525 & 0.990 & 0.993 & 0.993 & 0.992 \\
page-footer & 1094 & 1107 & 0.993 & 0.958 & 0.977 & 0.944 \\
page-header & 3 & 3 & 0.667 & 0.667 & 0.777 & 0.777 \\
section-header & 526 & 1513 & 0.989 & 0.987 & 0.992 & 0.987 \\
table & 491 & 532 & \emph{0.996} & \emph{0.998} & \emph{0.995} & \textbf{0.995} \\
table-of-contents & 73 & 74 & 0.973 & 0.959 & 0.979 & 0.979 \\
text & 575 & 2208 & 0.983 & 0.956 & 0.977 & 0.974 \\
\bottomrule
\end{tabular}
\end{table*}
We begin by comparing the layout detection performance of all models on the evaluation dataset. The models were evaluated using precision ($P$), recall ($R$), and mean average precision $\mathrm{mAP}$ at multiple (intersection-over-union) IoU thresholds. Specifically, following standard practices for an object detection task, $\mathrm{mAP@0.5}$ measures coarse localization accuracy, while $\mathrm{mAP@0.75}$ reflects stricter bounding box alignment. Additionally, we report COCO-style $\mathrm{mAP@0.5\!:\!0.95}$, which provides a comprehensive assessment across multiple overlap thresholds.

Table~\ref{tab:layout_metrics} summarizes the performance of various YOLO model variants on the evaluation set. The table shows that YOLO12x achieves the best overall results with the highest $\mathrm{mAP}$ ($\mathrm{mAP@0.5\!:\!0.95 = 0.9502}$), indicating a strong balance between precision and recall. Again, the YOLO12s model demonstrated the highest precision (0.9824), suggesting excellent accuracy in positive detections but slightly lower recall. Thus, the models in the YOLO12 family consistently outperformed the YOLO11 and YOLO26 variants across most metrics, highlighting improvements in both detection accuracy and consistency. While the YOLO26 and YOLO11  series achieved competitive results, particularly YOLO11m ($\mathrm{mAP@0.5\!:\!0.95}$ =0.9424), it slightly lagged behind the YOLO12 models. Thus, YOLO12x can be considered the most effective model for layout detection in terms of comprehensive performance across all evaluation criteria.

Based on the best-performing YOLO12x model, Table~\ref{tab:per_class_metrics} shows in the layout detection performance across layout units. The model demonstrated overall robust detection performance across all classes, except for images and, in particular, page headers. The slightly lower performance for the image class is due to the model occasionally confusing images with figures. Similarly, the human annotators also reported the same challenge during the annotation process. In contrast, the markedly lower performance for the page-header class suggests the presence of human annotation errors on the evaluation dataset.

\subsection{Layout Detection Performance Comparison with the Existing Methods}\label{experiments} 

Table~\ref{tab:layout_metrics_others} presents performance comparisons between the best-performing YOLO12 series and other prominent document layout detection methods. The results show that all YOLO12 models significantly outperform prior approaches, such as Surya-OCR, DocLayout (YOLO), Docling, and PaddleOCR, across all evaluation metrics. While earlier methods, including the YOLO-based DocLayout, achieved limited accuracy ($\mathrm{mAP@0.5\!:\!0.95 \leq 0.57}$) for the Khmer layout detection task, the YOLO12 models demonstrated substantial improvements, with YOLO12x achieving the highest overall performance ($\mathrm{mAP@0.5\!:\!0.95 = 0.9502}$) and YOLO12m attaining the best recall (0.9365). These findings highlight the limitations of the training data used by other methods, including the lack of Khmer layout data as well as scene document layout data. Similarly, the findings demonstrate the superior detection precision, robustness, and generalization capability of our models on the newly constructed dataset.

\begin{table*}[t]
\centering
\caption{Comparison of layout detection performance with the existing models on the evaluation set. \textbf{Bold}: \emph{best}. \textbf{Italic}:\emph{second best}.}
\label{tab:layout_metrics_others}
\begin{tabular}{lccccc}
\toprule
\textbf{Model} & \textbf{Precision} & \textbf{Recall} & \textbf{mAP@0.5} & \textbf{mAP@0.75} & \textbf{mAP@0.5:0.95} \\
\midrule
Surya-OCR~\cite{paruchuri2025surya} & 0.7186 & 0.6213 & 0.6742 & 0.5664 & 0.5156 \\
DocLayout (D4LA)~\cite{zhao2024doclayout} & 0.5205 & 0.3549 & 0.4318 & 0.2669 & 0.2859 \\
DocLayout (DocLayNet)~\cite{zhao2024doclayout} & 0.4323 & 0.1676 & 0.2876 & 0.2090 & 0.2012 \\
Docling~\cite{Docling} & 0.6477 & 0.6219 & 0.6520 & 0.5581 & 0.5094 \\
PaddleOCR-L~\cite{cui2025paddleocr30technicalreport} & 0.5988 & 0.6486 & 0.6431 & 0.5113 & 0.4914 \\
PaddleOCR-V2~\cite{cui2025paddleocr30technicalreport} & 0.6275 & 0.6837 & 0.6841 & 0.6287 & 0.5720 \\
\midrule
YOLO12s & \textbf{0.9843} & 0.9213 & \emph{0.9603} & 0.9568 & 0.9439 \\  
YOLO12m & \emph{0.9783} & \textbf{0.9365} & \textbf{0.9621} & \textbf{0.9588} & 0.9440 \\
YOLO12l & 0.9758 & 0.9227 & 0.9588 & 0.9565 & \emph{0.9479} \\
YOLO12x & 0.9528 & \emph{0.9364} & 0.9594 & \emph{0.9582} & \textbf{0.9502} \\

\bottomrule
\end{tabular}
\end{table*}

\subsection{Qualitative Evaluation}
\begin{figure}
\centering

\begin{subfigure}[b]{\textwidth}
    \centering
    {\includegraphics[width=\textwidth]{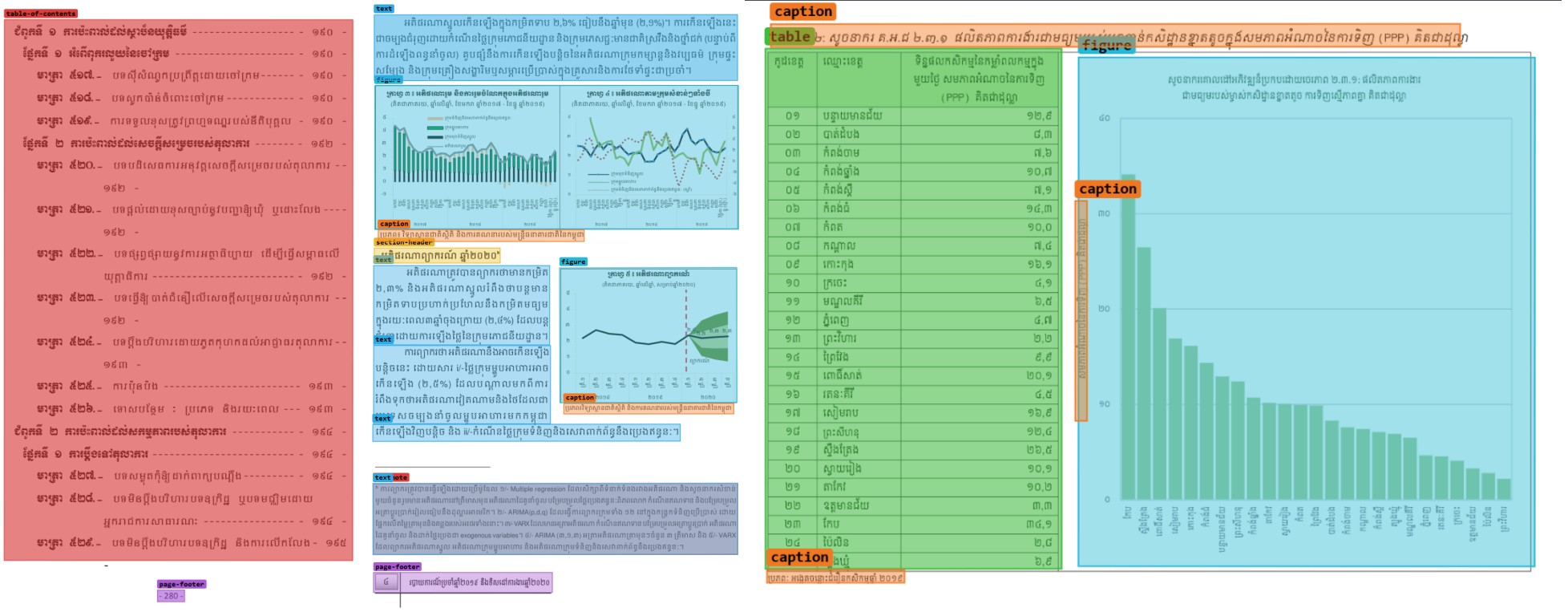}}
    \caption{PDF document case}

\end{subfigure}
\hfill
\begin{subfigure}[b]{\textwidth}
    \centering
    {\includegraphics[width=\textwidth]{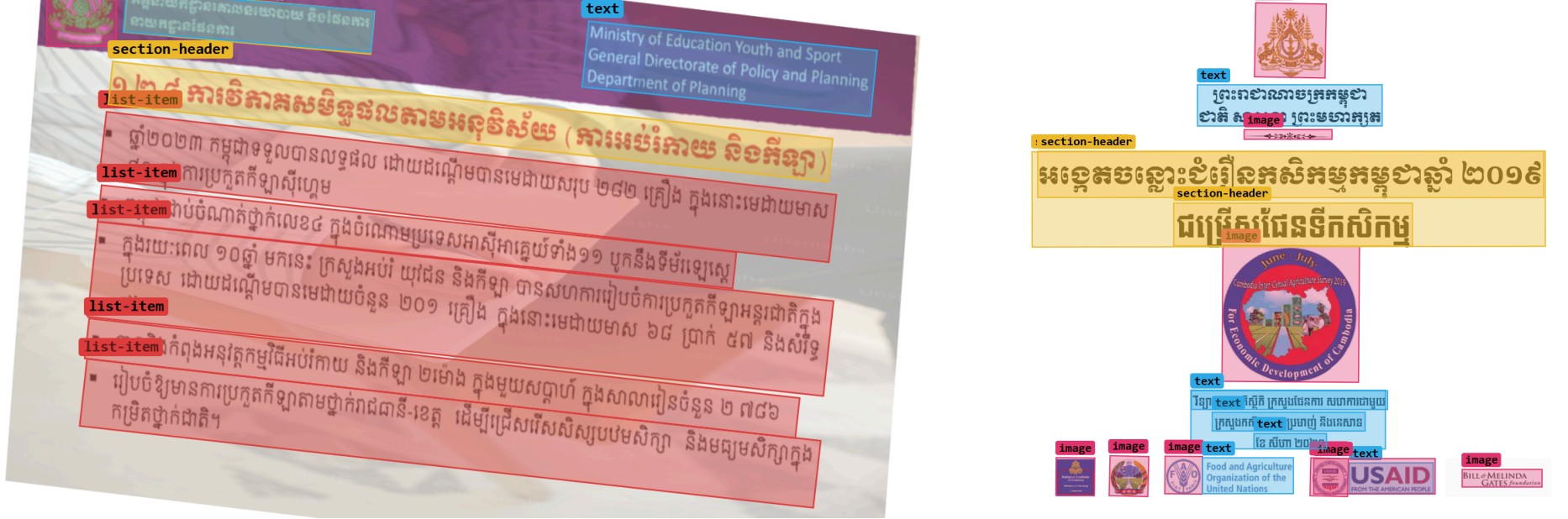}}
    \caption{Presentation slide case (with perspective deformations).}

\end{subfigure}
\hfill
\begin{subfigure}[b]{\textwidth}
    \centering
    {\includegraphics[width=\textwidth]{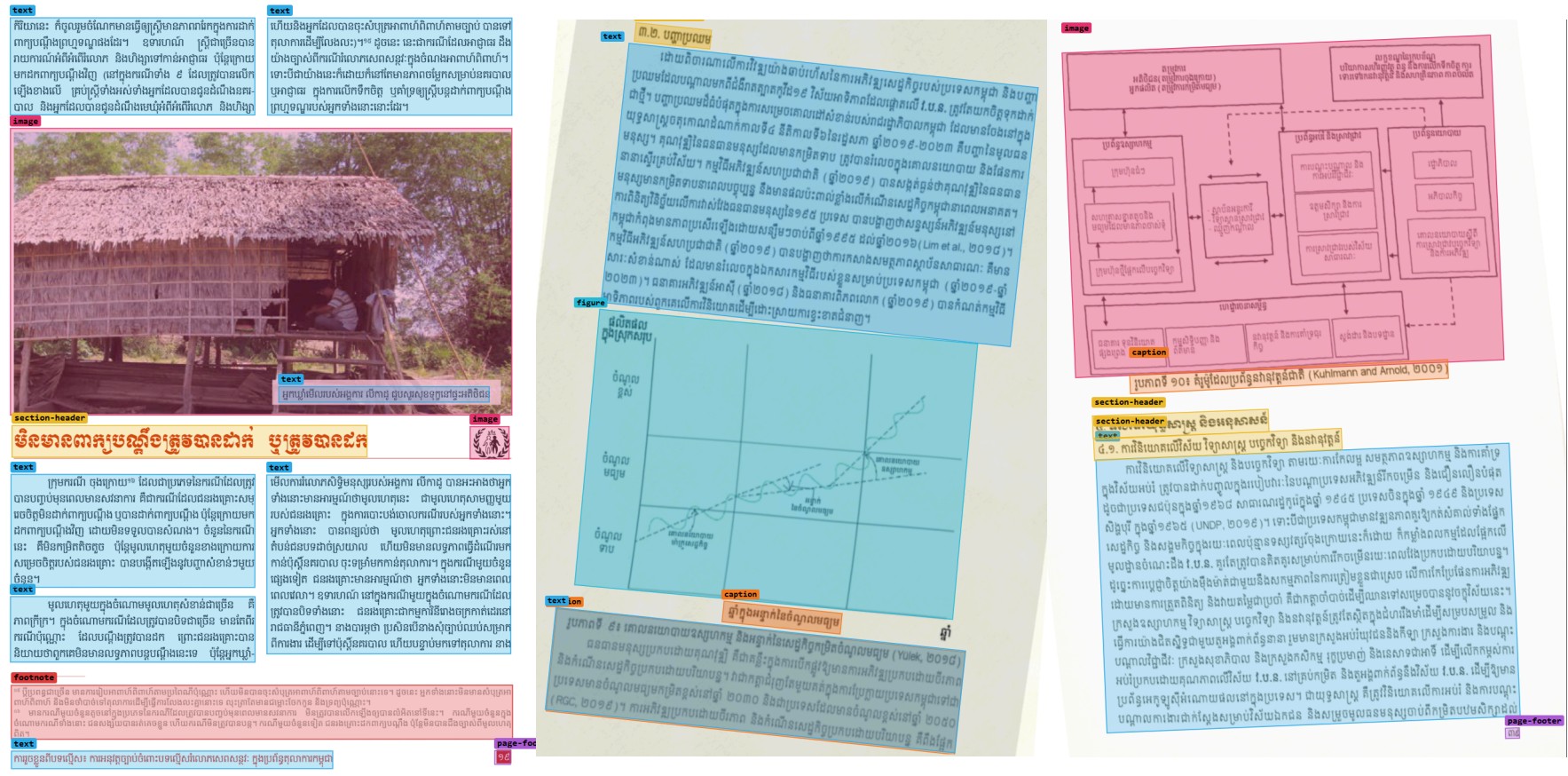}}
    \caption{Multi-column and scene document case}

\end{subfigure}
\caption{Some qualitative evaluation casess.  } \label{qual_eval}
\end{figure}

Figure~\ref{qual_eval} presents qualitative evaluation examples from the best-performing YOLO12x model. The examples include standard PDF documents with tables, figures, and tables of contents; slide documents with deformations; multi-column documents; and scene documents. Upon careful inspection, the model accurately detected the semantic layout units in all cases, regardless of transformations or deformations. 

Specifically, for PDF documents, the model correctly assigned semantic labels, including figures and their captions (even vertical captions). For scene documents, the model produced oriented bounding boxes that accurately enclosed the semantic layout units, including complex deformed figures with captions. These results further demonstrate the robustness of the baseline models for Khmer scene document layout detection.

\begin{figure}[t]
\centering

\centering
    {\includegraphics[width=\hsize]{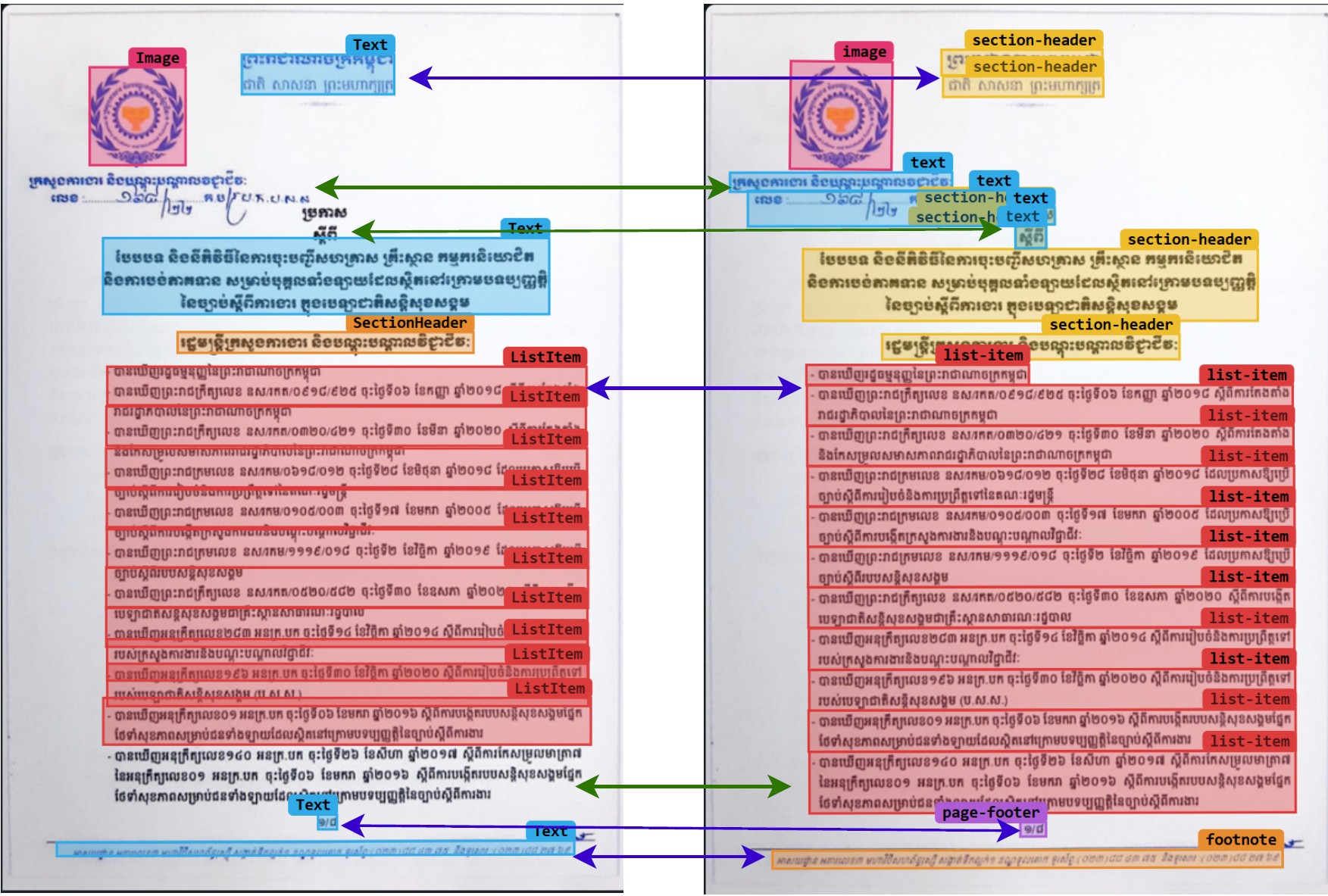}}
    \caption{Side-by-side comparison of Surya-OCR’s detection results (left) and those of our best-performing model (right). Blue arrow: wrong layout label. Green arrow: failed or missed detection. }

\label{sidebyside}
\end{figure}

In addition, Figure~\ref{sidebyside} compares the multilingual Surya-OCR’s detection results (left) and those of our best-performing YOLO12x model (right) for a sample input image. As shown in the figure, our best-performing model produces more accurate and semantically consistent layout detection results. Surya-OCR frequently assigns incorrect layout units (blue arrows) and fails to detect several text regions (green arrows), particularly in dense text regions or complex layout units. On the other hand, our model demonstrates robustness in identifying fine-grained layout elements, such as section headers, list items, and footnote.

\section{Limitations and Future Work}

We identify the following limitations associated with this study and future directions as follows:

\begin{enumerate}
\item The number of code-block, equation-block, and form instances in the current dataset remains limited. Therefore, future research should include documents related to mathematics or programming.

\item The current annotation scheme does not support nested layout units (e.g., textlines within a text block or rows and columns within a table). Therefore, future research should focus on developing annotation schemes that allow nested layout structures.

\item The current baseline models are based on the YOLO11, YOLO12, and YOLO26 series. Future research should focus on developing Khmer-specific layout detection models that are optimized for accurately understanding Khmer script and document layouts.
\end{enumerate}

\section{Conclusions}

In this paper, we present the largest single Khmer scene document layout dataset for training and evaluation. The dataset includes high-quality, human-curated annotations and scene document layout augmentations, making it suitable for real-world applications. To demonstrate its utility, we fine-tune the first Khmer layout detection baselines using YOLO-based architectures. Both the dataset and the baseline models represent key contributions to the Khmer document analysis and recognition community.
%
%
%
%

\end{document}